\documentclass[10pt,twocolumn,letterpaper]{article}

\usepackage{cvpr}
\usepackage{times}
\usepackage{epsfig}
\usepackage{graphicx}
\usepackage{amsmath}
\usepackage{amssymb}

\usepackage{multirow}

\usepackage{subfigure}
\usepackage{caption}
\usepackage{float}
\usepackage{authblk}

\usepackage[pagebackref=true,breaklinks=true,letterpaper=true,colorlinks,bookmarks=false]{hyperref}

 \cvprfinalcopy 

\def\cvprPaperID{4659} 

\ifcvprfinal\fi
\begin{document}

\title{Unsupervised Learning of Intrinsic Structural Representation Points}

\makeatletter 
\renewcommand\AB@affilsepx{\quad \protect\Affilfont} 
\makeatother

\author[1]{Nenglun Chen}
\author[2]{Lingjie Liu}
\author[1]{Zhiming Cui}
\author[1]{Runnan Chen}
\author[3]{Duygu Ceylan}
\author[4]{\\Changhe Tu}
\author[1]{Wenping Wang}
\affil[1]{The University of Hong Kong}
\makeatletter 
\renewcommand\AB@affilsepx{\\ \protect\Affilfont} 
\makeatother
\affil[2]{Max Planck Institute for Informatics}
\makeatletter 
\renewcommand\AB@affilsepx{\quad \protect\Affilfont} 
\makeatother
\affil[3]{Adobe Research}
\affil[4]{Shandong University}
\renewcommand\Authands{, }



\maketitle

\begin{abstract}
  Learning structures of 3D shapes is a fundamental problem in the field of computer graphics and geometry processing. We present a simple yet interpretable unsupervised method for learning a new structural representation in the form of 3D structure points. The 3D structure points produced by our method encode the shape structure intrinsically and exhibit semantic consistency across all the shape instances with similar structures. This is a challenging goal that has not fully been achieved by other methods. Specifically, our method takes a 3D point cloud as input and encodes it as a set of local features. The local features are then passed through a novel point integration module to produce a set of 3D structure points. The chamfer distance is used as reconstruction loss to ensure the structure points lie close to the input point cloud. 
Extensive experiments have shown that our method outperforms the state-of-the-art on the semantic shape correspondence task and achieves comparable performance with the state-of-the-art on the segmentation label transfer task. Moreover, the PCA based shape embedding built upon consistent structure points demonstrates good performance in preserving the shape structures. Code is available at \url{https://github.com/NolenChen/3DStructurePoints}
\end{abstract}

\section{Introduction}

Analyzing 3D shape structures is a fundamental problem in the field of computer graphics and geometry processing. One common way is to co-analyze a large collection of shapes, such as shape co-segmentation \cite{chen2019bae,zhu2019cosegnet}, shape correspondence estimation \cite{huang2014functional}, shape abstraction \cite{tulsiani2017learning,li2019supervised,sun2019abstraction} and 3D keypoint discovery \cite{suwajanakorn2018discovery}. An important requirement for such co-analysis is to leverage the semantic consistency among different shapes to discover semantically consistent features or structures that can facilitate different tasks. Quite a few structural representations have been proposed for 3D shapes, such as medial axis, curve skeleton and keypoints, which are designed for different tasks. Early works mainly use hand-crafted features and formulate it as optimization problems. While, they usually rely on parameter tuning, and are designed for specific tasks or datasets. Recently, deep learning techniques have been emerged to address these problems\cite{huang2018learning,muralikrishnan2019shape}. In this work, we propose a method to learn a new structural representation for establishing semantic correspondence for 3D point clouds.

\begin{figure}[h]
\centering
  \includegraphics[width=0.47\textwidth]{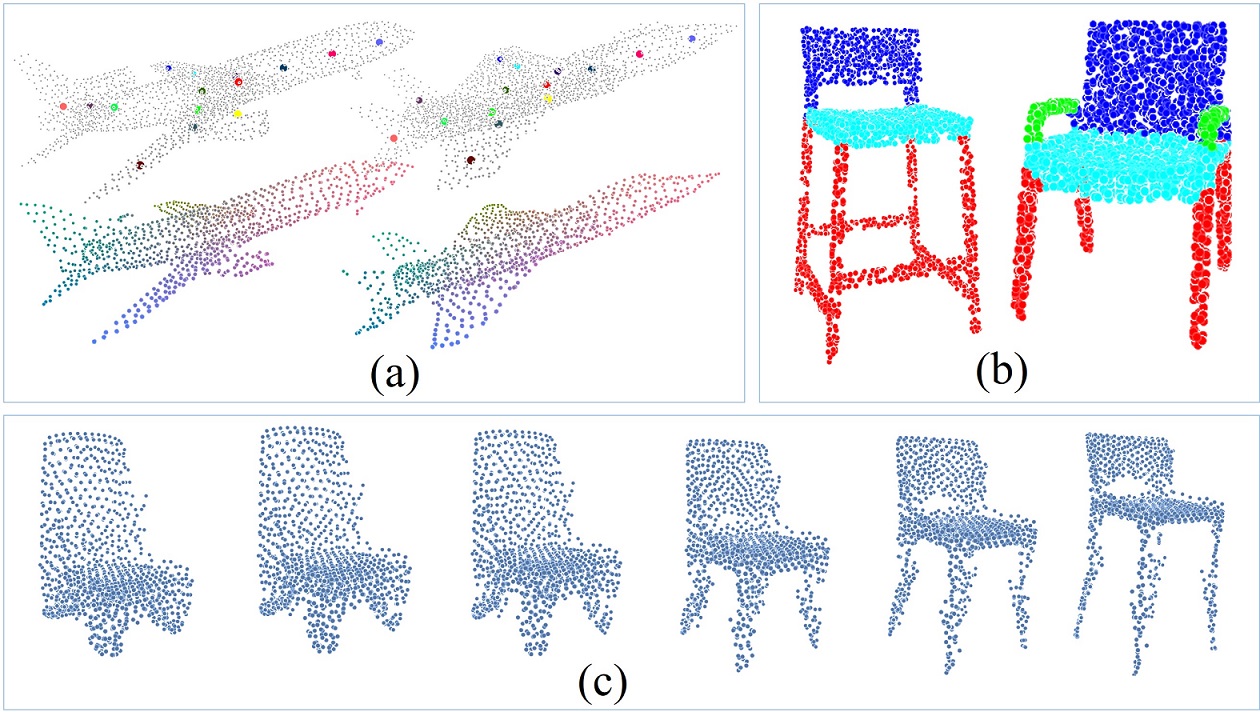}
      \vspace{-2mm}
  \caption{(a) Semantic correspondence. (b) Segmentation label transfer. (c) Visualization of one principal component of PCA embedding space.  }
  \label{fig:teaser}
  \vspace{-2mm}
\end{figure}

There has recently been active research on deep learning techniques for processing 3D point clouds. PointNet~\cite{qi2017PointNet1} is a pioneer work in this direction. After that, a number of network structures have been proposed and have achieved great success in various tasks. Most existing works mainly focus on generating accurate reconstruction of the target shapes, but the generated point clouds lack of structure information. Recently, several works \cite{tchapmi2019topnet,zhao20193d,yang2019pointflow,deprelle2019learning,shu20193d} have been proposed to generate structured point clouds, but with no guarantee for point-wise correspondence.

To address this issue, we propose a novel unsupervised method to learn semantically consistent structure points for 3D shapes in the same category. Given a 3D shape represented as a point cloud, the structure points generated by our network are a set of ordered points on the surface of the shape that provide a good abstraction and approximation of the input shape. Our architecture is simple but  directly interpretable. Extensive experiments show that our method is robust to different point samplings and can be generalized to real scanned point clouds unseen in the training stage. Our method achieves state-of-the-art performance on the semantic shape correspondence task and the segmentation label transfer task. Moreover, based on the high consistency of the structure points, we build a PCA based shape embedding that can preserve shape structures well. Figure \ref{fig:teaser} illustrates the related applications.

Our key contributions are summarized as follows:

\begin{itemize}
\item A simple yet interpretable unsupervised architecture for extracting semantically meaningful structure points for 3D shapes represented as point clouds. It is robust to different sampling of point clouds and is applicable to real scanned data.
\item Our method outperforms the state-of-the-art on the semantic shape correspondence task and achieves comparable performance with state-of-the-art methods on the segmentation label transfer task.
\item  A PCA based shape embedding built upon consistent structure points is able to preserve the shape structures well and has the potential to be used in several important tasks, such as shape reconstruction and shape completion.
\end{itemize} 
\section{Related Work}

\subsection{Shape Structure Analysis}

Recently, quite a few works have been proposed for learning keypoints as structural shape representations. Several unsupervised methods have been proposed to learn keypoints in 2D image domain. \cite{jakab2018conditional,lorenz2019unsupervised,zhang2018unsupervised}  disentangle the structure and appearance of 2D images for keypoints discovery.  For keypoint detection on 3D shapes, KeyPointNet \cite{suwajanakorn2018discovery} utilizes multi-view consistency to discover a sparse set of geometrically and semantically consistent keypoints across different shapes in the same category.

Unlike previous works, we conduct unsupervised learning of either sparse or dense consistent structure points as structural representation directly on 3D point clouds. Our method can be easily generalized to real scanned data.

\subsection{Deep Learning on Point Clouds}
Recent advances in deep neural networks have shown its great success in processing point clouds. PointNet \cite{qi2017PointNet1}, a pioneering work, utilizes point-wise MLP layers together with symmetric and permutation-invariant functions to aggregate information from all the points. To aggregate both local and global information, PointNet++ \cite{qi2017PointNet++} introduces a hierarchical structure that applies PointNet recursively on the partitioned point cloud. So-Net \cite{li2018so} uses Self-Organizing Map to model the spatial distribution of a point cloud to extract hierarchical features. Inspired by 2D convolution, \cite{simonovsky2017dynamic,wu2019pointconv,wang2018dynamic} reformulate the convolution operation and adapt it to point clouds. Our approach is built upon PointNet++ and it can also be built on other point cloud processing networks. 

Due to the irregular structures of point clouds, designing a decoder for generating point clouds is more difficult than that for 2D images. Previous works such as \cite{achlioptas2017learning,fan2017point,yu2018pu} mainly use MLP to generate point clouds from the encoded embedding. The point clouds generated by these methods lack structure information. FoldingNet \cite{yang2018foldingnet} and its variants \cite{deng2018ppf,deprelle2019learning,groueix2018atlasnet,yang2019pointflow} propose to generate the point clouds by deforming primitive structures.
\cite{lin2018learning} proposes a structural point cloud generator based on predicting structures at multiple viewpoints. \cite{shu20193d,tchapmi2019topnet} use tree structure to design decoder network for generating structured point clouds. \cite{zhao20193d} introduced capsule network for processing point clouds, in which shape structure information is encoded in the latent capsules. Our network, as a new kind of structured point cloud auto-encoder, produces structure points that exhibit point-wise consistency. Moreover, our PCA based structure points embedding also has the potential to be used in some important tasks like shape reconstruction and completion.

\begin{figure*}[ht]
\centering
  \includegraphics[width=0.95\textwidth]{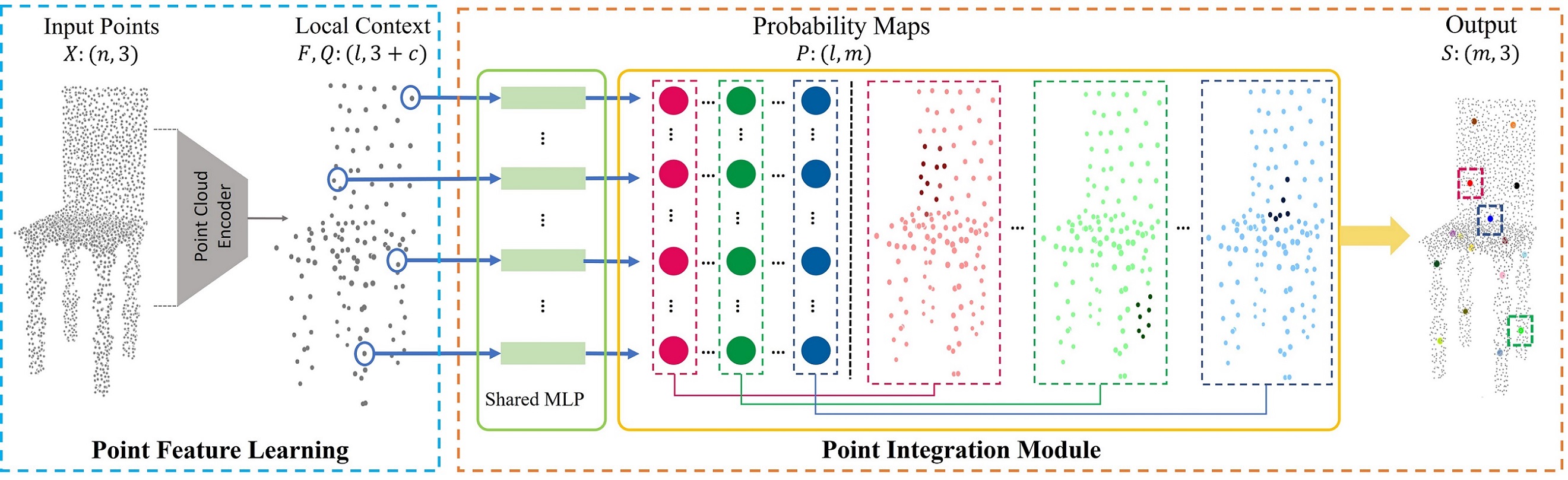}
      \vspace{-2mm}
  \caption{Pipeline of our network: Given a point cloud, PointNet++ is first used to extract the local contextual features $F$ and their corresponding points $Q$. Then, the features $F$ and the sample points $Q$ are further passed through the point integration module to produce the structure points $S$.}
  \label{fig:pipeline}
  \vspace{-4mm}
\end{figure*}

\subsection{3D Shape Correspondence}
Computing correspondences for 3D shapes is a fundamental task in geometric processings. Earlier methods \cite{huang2013fine,kim2013learning,kim2012exploring,rodola2014dense,xu2016data} mainly rely on hand-crafted descriptors or optimization among a collection of shapes to compute the correspondence. The functional map and its variants \cite{huang2014functional,ovsjanikov2012functional} provide a framework for representing maps between spaces of real-valued functions defined on shapes. 

Recently, deep learning techniques have been widely used in learning correspondence for 3D shapes. A series of methods \cite{chen2019edgenet,FCGF2019,gojcic2019perfect,huang2018learning} have been proposed to learn local descriptors for registrating point clouds. ShapeUnicode \cite{muralikrishnan2019shape} proposes a way to learn unified embedding for 3D shapes in different representations, and demonstrates its ability in establishing correspondence among 3D shapes. However, these methods require labeled pairwise correspondences in the training stage. Deep functional map \cite{sung2018deep} aims at building functional bases with deep learning, while indicator functions are still needed in the training stage. \cite{deng2018ppf,zeng20173dmatch} use self supervised learning to learn local descriptors about point clouds, and demonstrate good performance in registration of 3D scans.

Unlike previous methods, our approach learns consistent structure points across all shapes in a category in an unsupervised way. Based on the high consistency of the structure points, we can achieve good performance in transferring segmentation labels, and build the embedding space for point clouds directly with PCA.

\section{Method}

In this section, we introduce our end-to-end framework for learning intrinsic structure points from point clouds without explicit supervision. As illustrated in Figure \ref{fig:pipeline}, given a point cloud, $X=\{x_1, x_2,..., x_n\}$ with $x_i \in \mathbb{R}^3$, our model predicts an ordered list of structure points, $S=\{s_1, s_2,..., s_m\}$ with $s_i \in \mathbb{R}^3$. 
The network is trained for a collection of 3D shapes in the same category in an unsupervised manner. 
In the following, we first describe our network architecture which is composed of a PointNet++ encoder and a point integration module. Then we introduce a reconstruction loss for training. Finally, we show that the produced structure points exhibit semantic consistency across all the shapes in the same category, which is an essential property for shape co-analysis.

\subsection{Network Architecture}
Our network architecture is summarized in Figure \ref{fig:pipeline}. It can be seen as an encoder-decoder based architecture consisting of two modules, PointNet++ encoder and Point integration module. In the following, we provide more details.

\paragraph{PointNet++ Encoder}
Given a point cloud, $X=\{x_1, x_2,..., x_n\}$ with $x_i \in \mathbb{R}^3$, the first step is to extract sample points $Q=\{q_1, q_2,..., q_l\}(q_i \in \mathbb{R}^3$) with the local contextual features $F=\{f_1,f_2,...,f_l\}(f_i \in \mathbb{R}^c$), where $l$ indicates the number of sample points and $c$ is the dimension of the feature representation.

We build our architecture upon the PointNet++ \cite{qi2017PointNet++} encoder, which extracts point cloud features by adaptively combining features from multiple scales. The PointNet++ encoder is composed of multiple set abstraction levels. Each level consists of three key layers: sampling layer, multi-scale grouping layer and PointNet layer for processing and abstracting the input points in a hierarchical way. We refer the reader to the original paper for more details.

\begin{figure}[h]
\centering
  \includegraphics[width=0.45\textwidth]{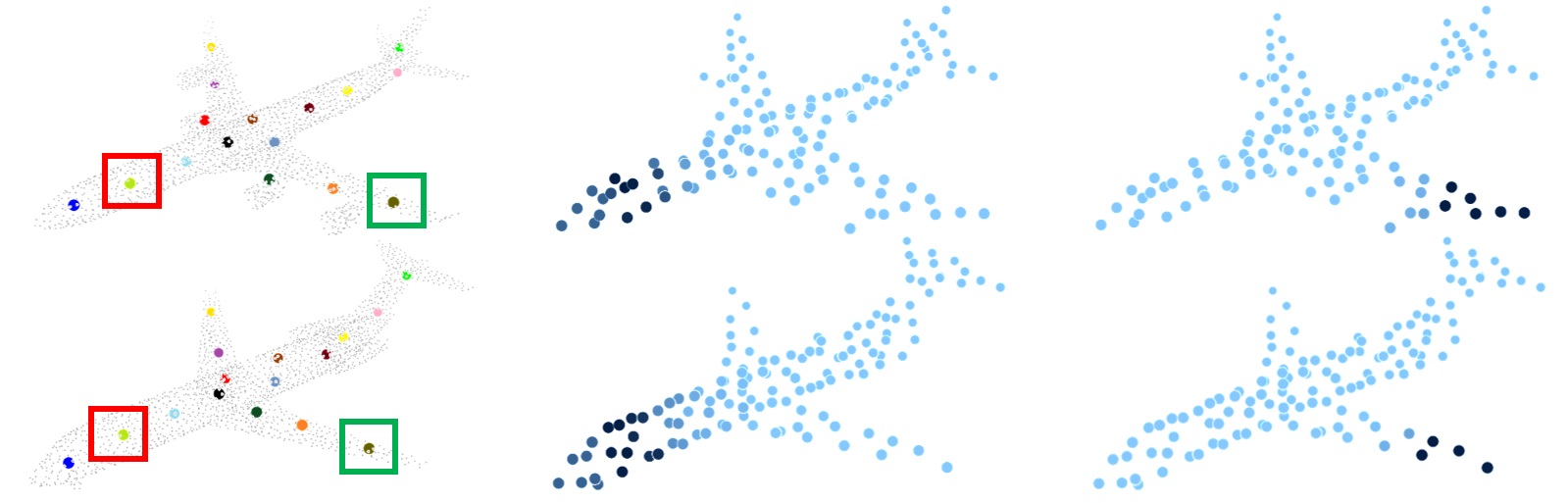}
      \vspace{-2mm}
  \caption{The probability maps of two structure points (in red and green boxes) are shown in
  second and third columns. Darker color indicates higher probability.}
  \label{fig:feat_prob_map}
    \vspace{-5mm}
\end{figure}

\paragraph{Point Integration Module}
The input to this module is the sample points $Q$ with their local contextual features $F$ of size $l \times (3+c)$ obtained by the encoder. The outputs are $m$ structure points $S$. Detailly, given the local contextual features $F$, we first apply a shared multi-layer perceptron (MLP) block followed by
softmax as activation function to produce the probability maps $P=\{p_1, p_2, ..., p_m\}$ . 
The element $p_i^j$ in the probability map $p_i$ indicates the probability of the point $q_j$ being the structure point $s_i$. 
Therefore, the structure points $S$ can be computed as follows:

\begin{equation}\label{eq:integration}
s_i = \sum_{j=1}^{l} q_j p_i^j   \quad \textrm{with} \quad \sum_{j=1}^{l} p_i^j=1  \quad \textrm{for each}  \quad i
\end{equation}
Equation \ref{eq:integration} is a convex combination of the points $Q$, ensuring that the predicted structure points $S$ are located within the convex hull of the points $Q$.

In Figure \ref{fig:feat_prob_map}, we visualize the learned probability map for different structure points. The first column shows the input point clouds (in grey) and their corresponding 16 structure points (in color). The probability map of two structure points marked with red and green boxes are illustrated in the second and third columns. Note that, the learned probability maps have concentration effect, though we do not explicitly enforce them to be concentrated in local regions.

\subsection{Reconstruction Loss}
To train the network in an unsupervised manner, we define our reconstruction loss based on the Chamfer distance (CD)~\cite{fan2017point} to constrain the predicted structure points to be close to the input point cloud. Specifically, the reconstruction loss is the Chamfer distance between the predicted structure points $S$ and the input points $X$:

\begin{equation}\label{eq:general_cd}
L_{rec}(S, X)=  \sum_{s_i \in S} \min_{x_j \in X} \|s_i-x_j\|_2^2 + \sum_{x_j \in X} \min_{s_i \in S} \|s_i-x_j\|_2^2
\end{equation}

\subsection{Cross-Object Consistency}
\begin{figure*}[ht]
\centering
  \includegraphics[width=0.97\textwidth]{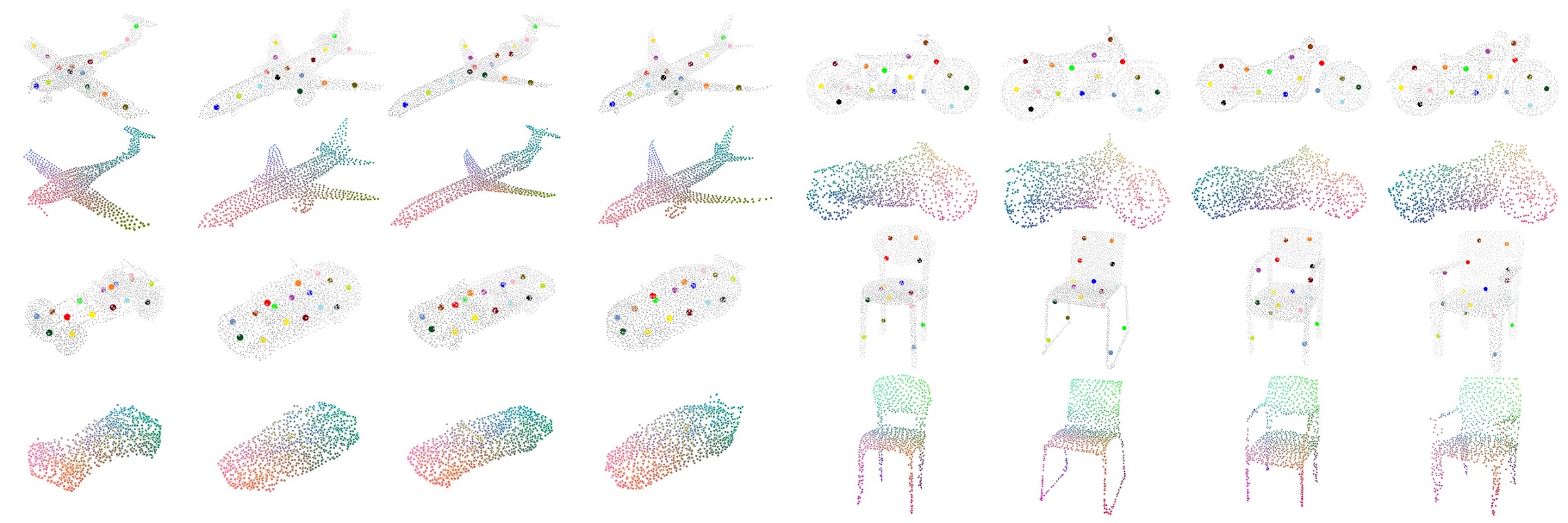}
      \vspace{-2mm}
  \caption{Consistency across different shapes. The produced 16 structure points (colored) for different input point clouds (grey) are shown in 1st and 3rd rows, and the produced 1024 structure points are in 2nd and 4th rows. Corresponding structure points are in the same color.}
  \label{fig:fpts_consis}
    \vspace{-4mm}
\end{figure*}

We do not explicitly enforce the structure points to be consistent for different instances, but the network is able to generate semantically consistent probability maps $P$ for the objects in the same category, leading to the consistency of the produced structure points as shown in Figure \ref{fig:fpts_consis}. In the following, we give a detailed explanation about the consistency of structure points.

To simplify discussion, we first consider two identical shapes $A$ and $B$ defined by the same point cloud $X$, except that these points are processed in possibly different orders for $A$ and $B$.
After passing through PointNet++ feature extractor and the shared MLP, a set of sample points of shape $A$ are extracted with the corresponding $m$-dimensional probability vectors.
Let us denote these sample points of $A$ by the ordered sequence of points $(a_1, a_2, ..., a_{\ell})$, with the corresponding probability vectors $\zeta_i = (p^i_1, p^i_2, ..., p^i_m)$, $i=1, 2, \dots, \ell$. Since $B$ is identical to $A$, we may suppose that the sample points  $(b_1, b_2, ..., b_{\ell})$ obtained in the same way for $B$ are a permutation of the sample points $(a_1, a_2, ..., a_{\ell} )$ for $A$. So we may denote $b_i = a_{\pi(i)}$, where $\pi (.)$ is a permutation function of $\{1, 2, .., \ell\}$. Then the probability vectors of the sample points $b_i$ of shape $B$ are $\zeta_{\pi (i)} = (p^{\pi (i)}_1, p^{\pi (i)}_2, ... p^{\pi (i)}_m)$, $i=1, 2, \dots, \ell$. And the $j$-th structure point for shape $A$ is given by: 
\begin{equation}
s^A_j = \sum_{i=1}^l p^i_j a_i
\end{equation}
Then for the $j$-th structure point $s^B_j$ for shape $B$, we have:
\begin{equation}
 s^B_j = \sum_{i=1}^\ell p^{\pi(i)}_j b_i = \sum_{i=1}^\ell p^{\pi(i)}_j a_{\pi(i)} =  \sum_{i=1}^\ell p^i_j a_i = s^A_j
\end{equation}
That is, the $j$-th structure point for shape $A$ is the same as the $j$-th structure point for shape $B$. When two shapes $A$ and $B$ are similar, because the neural networks in the pipeline are continuous mappings, the structure points $\{s^A_j\}_{j=1}^m$ of $A$ will be individually close to their corresponding structure points $\{s^B_j\}_{j=1}^m$ of $B$.

\section{Experiments}

\subsection{Datasets}

We conduct most of our experiments on the ShapeNet \cite{yi2016scalable} dataset. In the 3D semantic correspondence task, we use the same datasets in \cite{huang2018learning}, where ShapeNet \cite{yi2016scalable} and the BHCP benchmark \cite{kim2013learning} are used for training and testing respectively. For segmentation label transfer task, ShapeNet part dataset \cite{yi2016scalable} is used. Farthest point sampling \cite{eldar1997farthest} is applied to sample point clouds from 3D shapes.

\subsection{Implementation Details}
The PointNet++ encoder we used is composed of two set abstraction levels with 512 and 128 grouping centers respectively. Multi-scale grouping (MSG) is used to combine the multi-scale features. The MSG layers in two levels contain scales (0.1, 0.2, 0.4) and (0.2, 0.4, 0.8) respectively. The output of the PointNet++ encoder contains $l=128$ sample points, with each sample point having 640 dimensional local contextual features. The configuration of the MLP block in the point integration module depends on the number of structure points. Specifically, for $m=512$ structure points, the MLP block contains 3 layers with the neuron numbers (640, 512, 512). The dropout ratio is set to 0.2 to avoid over-fitting of the MLP block. Adam\cite{kingma2014adam} is used as the optimizer. We train our network on a single NVIDIA GTX 1080Ti GPU with less than 1 hour per category.

\subsection{Consistency Across Objects}
The structure points produced for different shapes are visualized in  Figure \ref{fig:fpts_consis}, where the input point clouds (grey) and the sparse structure points (colored) are shown in the 1st and 3rd rows, and the corresponding dense structure points are illustrated in the 2rd and 4th rows. Corresponding structure points are visualized in the same color. It can be seen that our approach can generate sparse and dense structure points in a consistent manner for the shapes with similar structures. Note that such correspondence may not exist in the regions with significant structure differences. An example is the chairs with and without armrest in Fig. \ref{fig:fpts_consis}.

\subsection{3D Semantic Correspondence }

To evaluate the cross-object consistency of the structure points, we test our approach on the task of computing 3D shape semantic correspondence and compare it with the state-of-the-art methods that have been specifically tuned for this task including LMVCNN \cite{huang2018learning}, ShapeUnicode \cite{muralikrishnan2019shape} and EdgeNet \cite{chen2019edgenet}, as well as some point cloud encoder-decoder architectures: AtlasNet2 \cite{deprelle2019learning}, FoldingNet \cite{yang2018foldingnet}, TopNet \cite{tchapmi2019topnet} and 3DPointCapsule \cite{zhao20193d}. We also use the critical points in PointNet \cite{qi2017PointNet1} and the FCN based decoder described in \cite{achlioptas2017learning} as baseline comparisons. Since LMVCNN is rotation-invariant, for fair comparison with it, we train our method with rotation augmentation and test on the point clouds with arbitrary rotations.

Specifically, to make the network applicable to a rotated point cloud, we perform a PCA (Principle Components Analysis) \cite{jolliffe2011principal} based augmentation for training. We first compute three main axes for each shape in the training dataset. In each training iteration the main axes of each shape are randomly swapped and the shapes are aligned according to the swapped axes. The consistency of the structure points for original and augmented shape are enforced with Mean Square Error (MSE) loss.  During testing, we also compute the three main axes for each shape, and align it accordingly. Compared to data augmentation with random rotations, the PCA based scheme can reduce the rotation space, and makes the network converge more efficiently.

We use the datasets used in \cite{huang2018learning}, where ShapeNet \cite{yi2016scalable} and the BHCP benchmark \cite{kim2013learning} are used for training and testing respectively. Note that we use airplanes to train the network for helicopters in the same way it did in \cite{huang2018learning} since helicopters are not included in the training data.

\begin{figure}[htbp]

  \includegraphics[width=0.47\textwidth]{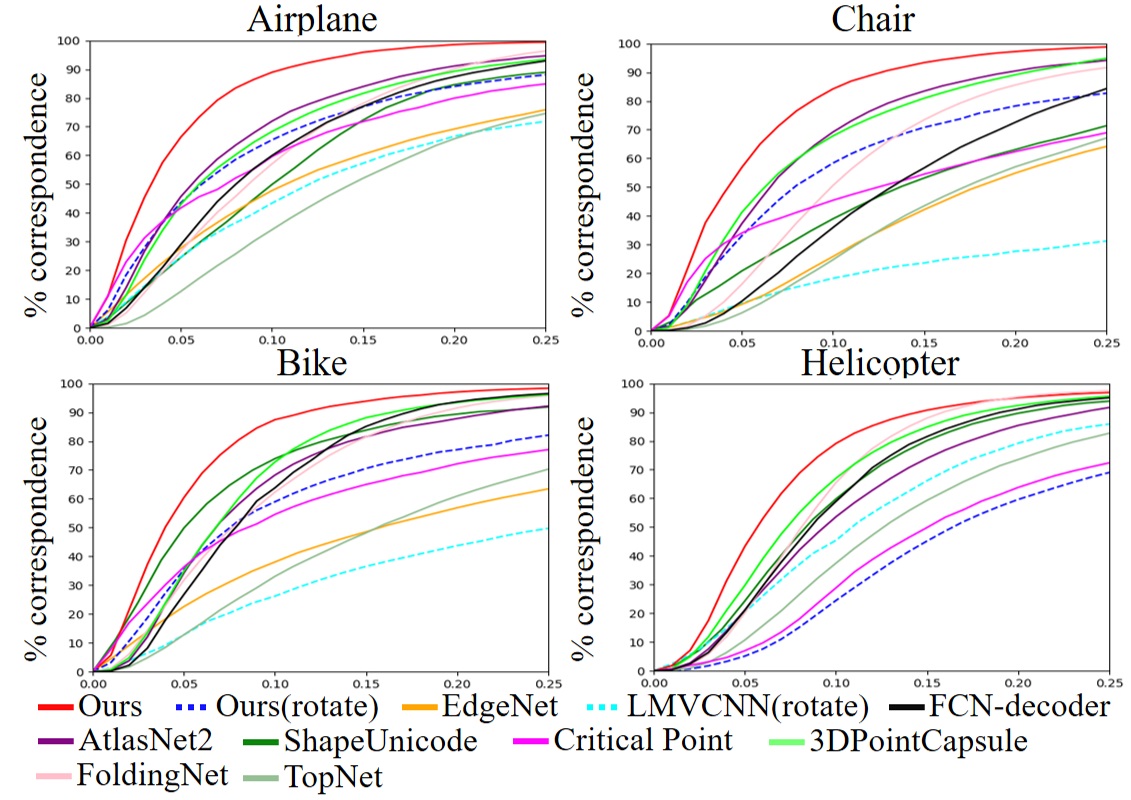}
      \vspace{-2mm}
  \caption{Correspondence accuracy for each category in the BHCP benchmark.}
  \label{fig:bhcp_ca}
    \vspace{-4mm}
\end{figure}

Our network is trained on a collection of aligned shapes represented by 2048 points with 512 structure points as output. During testing, given a point $x_q$ on a 3D shape, we first find its closest structure point $s_q$, and then use the corresponding structure point $s_q'$ on the target shape as the corresponding point of $x_q$. The correspondence accuracy is measured by the fraction of the correspondences that are correctly predicted with the error below given Euclidean thresholds.

In Figure \ref{fig:bhcp_ca}, the solid lines show the results tested on the aligned data and the dotted lines denote the results tested on the unaligned data. We can clearly see that our method significantly outperforms other state-of-the-art methods. Moreover, we demonstrate the good generalization to unseen categories (e.g. train with airplanes and test with helicopters). However, in this case, the performance of our method on the rotated data is not as good as the aligned one. That is because the PCA of helicopters is quite different from that of airplanes, therefore more difficult for the network to adapt to the unseen category of helicopters.

\subsection{Example based Label Transfer}

We further evaluate the quality of the structure points by transferring segmentation labels from few examples, and compare our average IOU with a state-of-the-art shape segmentation method BAE-NET ~\cite{chen2019bae} on shapenet part\cite{yi2016scalable} dataset. BAE-NET has different settings(e.g. unsupervised, one-shot and few-shot), we compare our method with it's few-shot setting. Specifically, our network is trained on the training set without any label. During testing, to label a point $x_q$ on a shape, we find its closest structure point $s_q$ as well as a list of corresponding structure points $E=\{s_q^1, s_q^2,...,s_q^k\}$ on $k$ example shapes, and the label of structure point $s_q' \in E$ which has the most similar features with $s_q$ is transferred to $x_q$.

\begin{figure}[htbp]
  \includegraphics[width=0.45\textwidth]{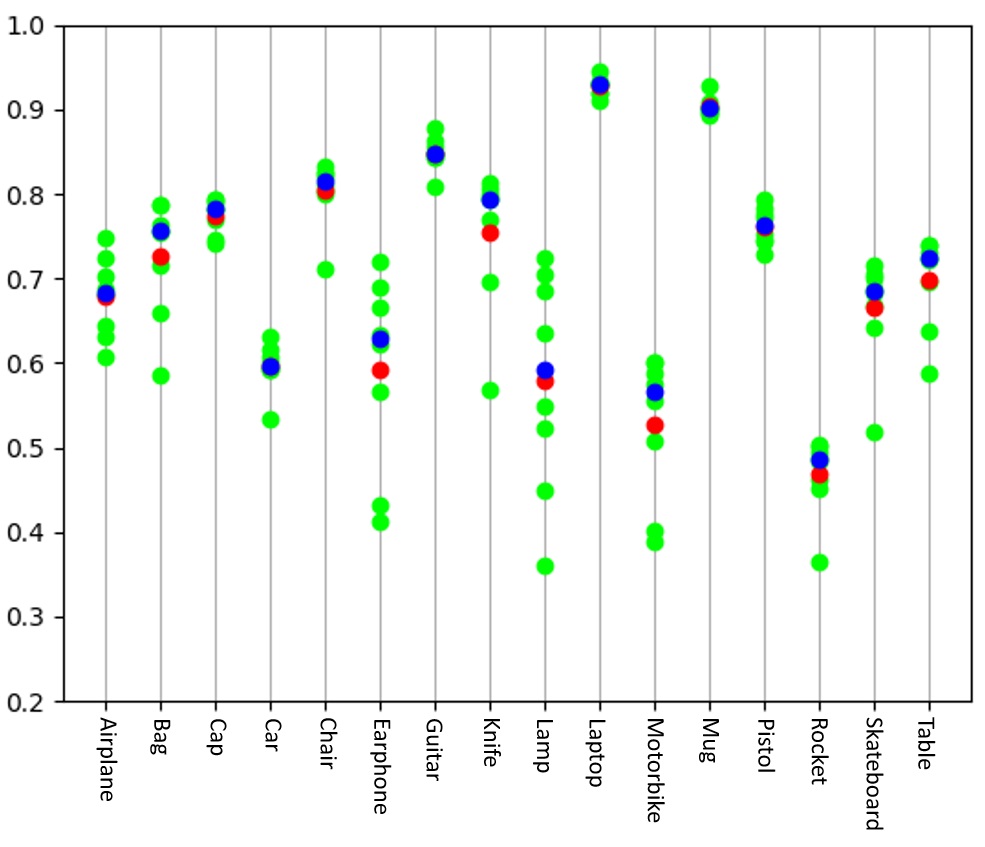}
      \vspace{-2mm}
  \caption{Label transfer results based on 3 examples. For each category, the average IOUs by 8 randomly selected example sets (green dots) are illustrated. The mean and median are shown in red and blue. }
  \label{fig:vis_seg_stats}
    \vspace{-2mm}
\end{figure}

\begin{table*}[h]
	\centering
	\begin{center}
	\scalebox{0.73}{
	\begin{tabular}{l c c c c c c c c c c c c c c c c}
		\hline
        \cline{1-17}
		~ & Airplane & Bag & Cap & Car & Chair & Earphone & Guitar & Knife & Lamp & Laptop & Motorbike & Mug & Pistol & Rocket & Skateboard & Table\\
		
		\cline{1-17}
		BAE-Net & 74.7 & \textbf{83.9} & \textbf{85.5} & - & \textbf{86.0} & \textbf{76.2} & 87.8 & \textbf{83.6} & 70.1 & \textbf{94.8} & \textbf{64.6} & \textbf{94.8} & 78.7 & \textbf{52.1} & \textbf{74.2} & 73.3\\
		\cline{1-17}
		Ours & \textbf{74.9} & 78.6 & 79.3 & \textbf{60.6} & 84.5 & 69.0 & \textbf{87.9} & 80.6 & \textbf{72.4} & 94.4 & 60.0 & 92.7 & \textbf{79.4} & 50.3 & 70.4 & \textbf{73.9}\\
		 \cline{1-17}
		\end{tabular}}
	\end{center}
	    \vspace{-4mm}
		\caption{Comparison of our label transfer results against BAE-NET with 3 labeled exemplars on the ShapeNet part dataset\cite{yi2016scalable} measured with average IOU(\%)}\label{tab:miou_label_transfer}
		    \vspace{-1mm}
\end{table*}

\begin{figure}[ht]
\centering
  \includegraphics[width=0.45\textwidth]{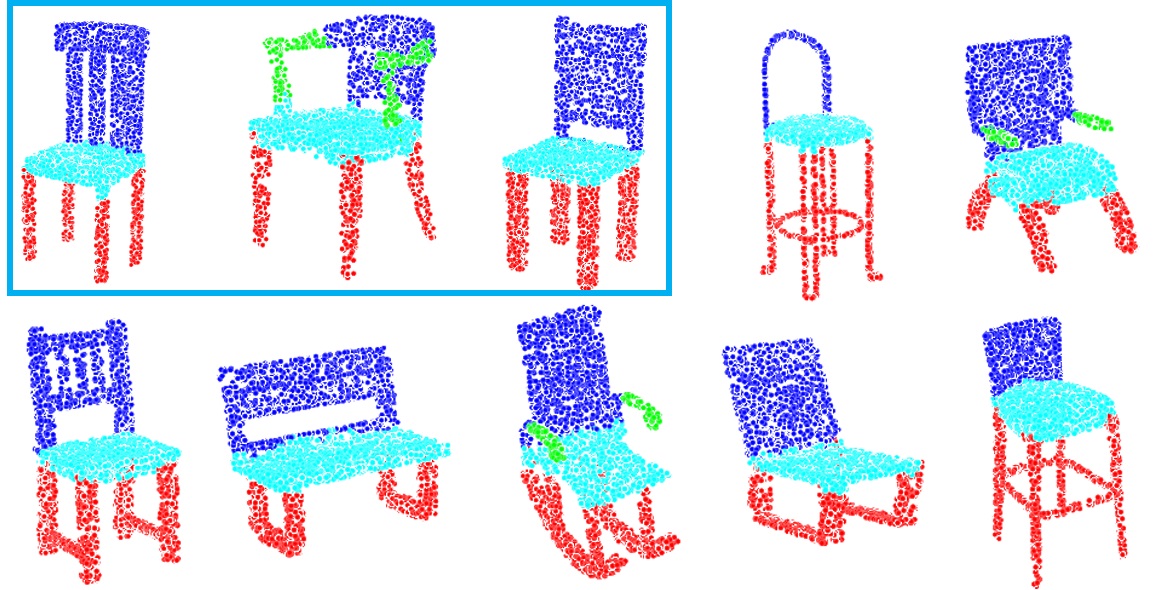}
      \vspace{-2mm}
  \caption{Qualitative results for label transfer, the example shapes are in the blue box.}
  \label{fig:vis_seg}
    \vspace{-4mm}
\end{figure}

As discussed in \cite{chen2019bae}, the accuracy of the segmentation results will be affected by the selection of the example shapes in the few-shot setting. Empirically, the exemplars should contain all the segmentation labels and represent the variations of the dataset. In Figure \ref{fig:vis_seg_stats}, we show all the results of our segmentation label transfer by using 8 randomly selected exemplar sets, each set contains 3 exemplars. For the categories with regular shapes (e.g. Laptop), the results are consistent. While, the accuracy may vary a lot for the categories with large shape variations (e.g. 'Earphone'). In Table \ref{tab:miou_label_transfer}, the best performing results are reported for both methods, and Figure \ref{fig:vis_seg} demonstrates some qualitative results. We can see that, our label transfer results are on par with BAE-Net in most categories with structured shapes. For some categories such as 'Bag', 'Cap', and 'Earphone', due to their large variations, 3 exemplars are not enough for our method. For the 'Car' category, since it contains flat surface that BAE-Net cannot separate, we do not report its result. Moreover, in BAE-NET, the segmentation labels are either pre-defined by the user (few-shot setting) or implicitly defined by the network (unsupervised setting); once the training is done, the labels cannot change. In contrast, our work achieves the few-shot label transfer by transferring segmentation labels directly from exemplars, thus has the potential of transferring arbitrary labels (e.g., labels with different hierarchies) after training. (More results are in the supplementary material)

\subsection{PCA based Shape Embedding}
\begin{figure*}[htbp]
\centering
  \includegraphics[width=0.95\textwidth]{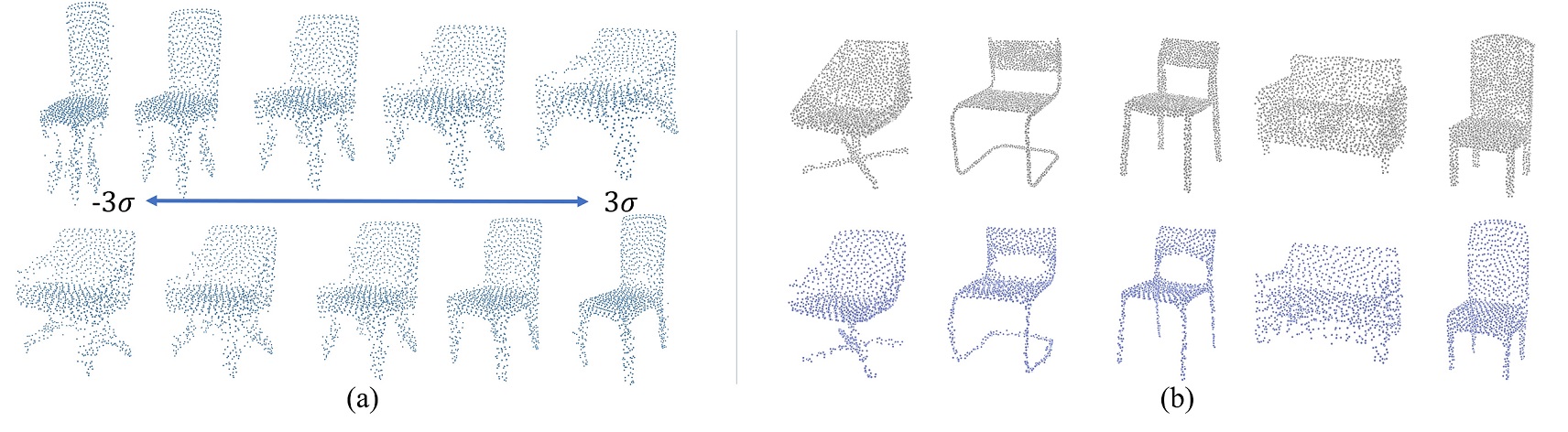}
     \vspace{-3mm}
  \caption{PCA based shape embedding. (a) Visualization of the PCA embedding space by adding the PCA principle components to the mean shape with ratio from $-3\sigma$ to $3\sigma$. The first two components are used in the 1st and 2rd row respectively (b) Input point clouds (first row) and the corresponding reconstructions (second row) with $50$ principal components.}
  \label{fig:vis_pca_rep}
    \vspace{-4mm}
\end{figure*}

Given high quality shape correspondences, we can build a shape embedding space based on PCA. Specifically, given a collection of point clouds, we produce $m$ structure points $S=\{s_1^1, s_1^2, s_1^3,...,s_m^1,s_m^2,s_m^3\}$ for each point cloud with the proposed network. Where $\{s_i^1,s_i^2,s_i^3\}\in \mathbb{R}^3$ denotes the location of $i$th structure point. Then a shape morphable model \cite{blanz1999morphable} can be constructed based on the learnt structure points. Thus, a new shape $X$ can be be represented as a linear combination of the principal components:

\begin{equation}\label{eq:morphable_model}
X = \bar{S} + \sum_{i=1}^{k}\alpha_i  c_i
\end{equation}

Where $\bar{S}$ is the mean shape of the structure points, $k$ is the number of principal components, $c_i$ and $\alpha_i$ denote the $i$th principal component and the corresponding coefficient respectively. Figure \ref{fig:vis_pca_rep}(a) shows the PCA embedding space by adding the first two PCA principal components to the mean shape with ratio from $-3\sigma$ to $3\sigma$, where $\sigma^2$ denotes the eigen value of the corresponding principal component. Figure \ref{fig:vis_pca_rep}(b) shows some reconstruction results with only $50$ principal components. We can see that the structures of the shapes can be well preserved with our PCA based embedding. (More results are in supplementary material)

\section{Ablations and Visualizations}

\subsection{Feature Embedding Visualization}
To better understand the mechanism of the proposed method, we extract and visualize the latent features learned by the network. The local contextual features $F$ produced by PointNet++ are first weighted by the probability map $P$ to get per-point features $H=\{h_1,h_2,...,h_m\}$ with $h_i \in \mathbb{R}^c$ for each structure point:

\begin{equation}\label{eq:feat_weight_sum}
h_i = \sum_{j=1}^l f_j p_i^j \quad \textrm{with} \quad \sum_{j=1}^{l} p_i^j=1  \quad \textrm{for each}  \quad i
\end{equation}

The per-point features $H$ for all the shapes in the ShapeNet testing dataset with the same category are then embedded in a 2-dimensional embedding space  using t-SNE \cite{maaten2008visualizing} for visualization.

\begin{figure}[htbp]
  \includegraphics[width=0.49\textwidth]{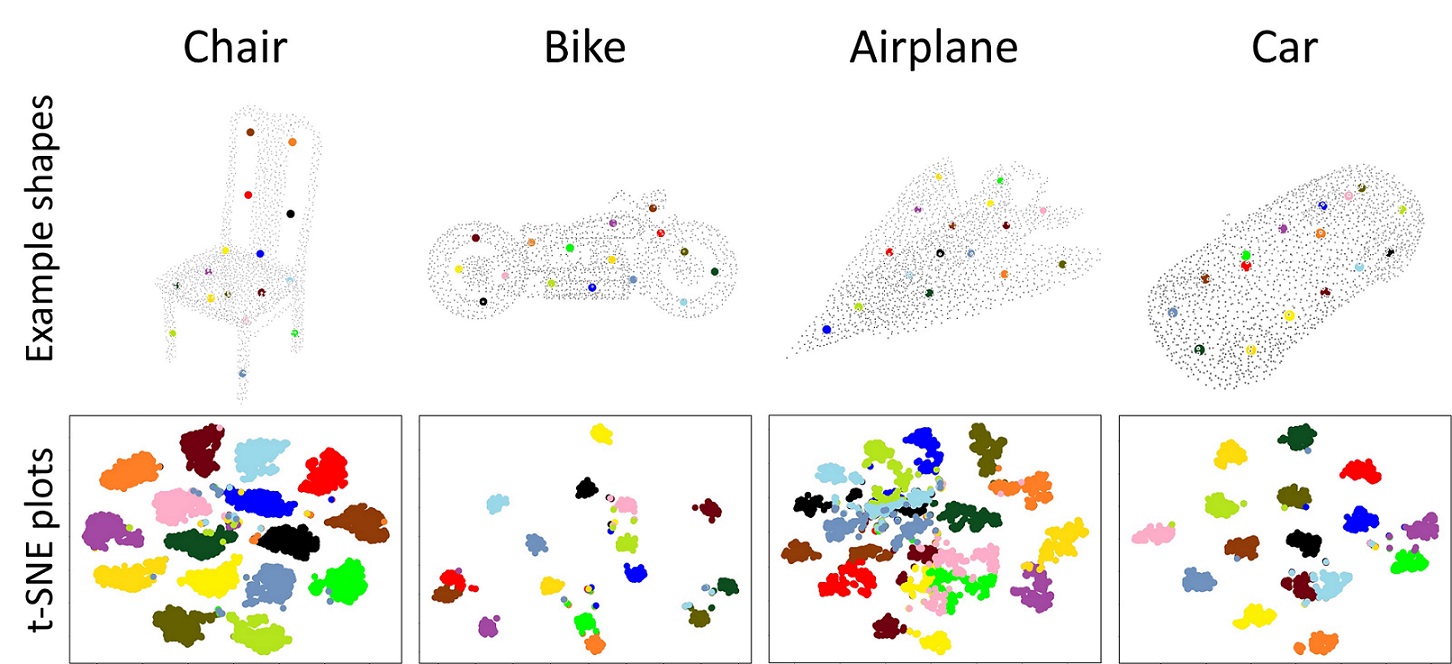}
      \vspace{-2mm}
  \caption{The T-SNE embedding of the learned features. The first row shows examples of structure points overlaid with the input point clouds, and the second row shows the corresponding t-SNE embeddings.}
  \label{fig:tsne}
    \vspace{-4mm}
\end{figure}

Figure \ref{fig:tsne} illustrates the t-SNE results of the embedding space computed on four categories when 16 structure points are predicted. The 2D points in the plot are colored with 16 unique colors each corresponding to a specific structure point. Note that, the learned features are well clustered, which means the structure points with the same semantic location tend to have similar features, though we do not explicitly enforce the consistency of the latent features. One may note that symmetry structure points on the same shape do not have similar features, this is because the PointNet++ encoder we used is not symmetric invariant. One may consider adding symmetry constraints into the loss or using a symmetry invariant feature encoder to make the structure points symmetric invariant.

\subsection{Robustness to Sampling Densities}

To evaluate the robustness of our approach to input point clouds with different densities, we train our network on 2048 points sampled on each shape, and test the network on input point clouds with different densities.

We use the point-wise average Euclidean distance to measure the stability of the produced structure points with different input densities in comparison with the structure points produced from 2048 input points. To generate non-uniformly sampled points, we first randomly sample a relatively small number of seed points from the initial point cloud, and remove points near the seeds with certain probability. This creates a set of non-uniformly distributed points with missing points (or ``holes'') around the seed points.

\begin{table}[h]
	\begin{center}
	\scalebox{0.88}{
	\begin{tabular}{l c c c c}
		\hline
		\multirow{2}*{Sample Num} & \multicolumn{3}{c}{Average Distance(\%)} \\
        \cline{2-5}
		~ & 256 & 512 & 1024 & 4096\\
		\cline{1-5}
		Chair & 0.7346 & 0.1308 & 0.0200 & 0.0027 \\
		Airplane & 0.1209  & 0.0169 & 0.0022 & 0.0001\\
		Car & 0.5470 & 0.1348 & 0.0099 & 0.0017\\
		Motorbike & 0.3070 & 0.0610  &  0.0195& 0.0111\\
		 \hline
	\end{tabular}}
	\end{center}
	    \vspace{-3mm}
	\caption{Stability of the network for producing 1024 structure points with input points of different uniform densities.}\label{tab:dif_sample_ave_dis}
    \vspace{-2mm}
\end{table}

\begin{figure}[htbp]
\centering
  \includegraphics[width=0.49\textwidth]{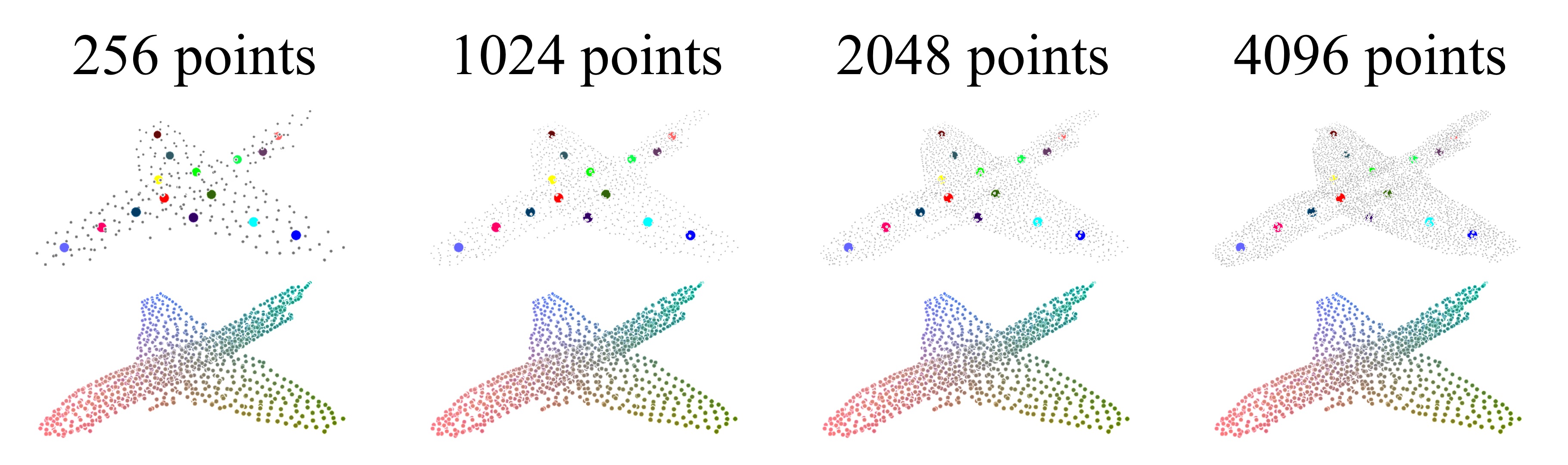}
      \vspace{-2mm}
  \caption{Robustness to different samplings. The 1st row shows 16 structure points and different numbers of input point clouds. The 2nd row shows 1024 structure points. Corresponding structure points have the same color.}
  \label{fig:diff_sample_vis}
    \vspace{-4mm}
\end{figure}

\begin{figure}[htbp]
\centering
  \includegraphics[width=0.47\textwidth]{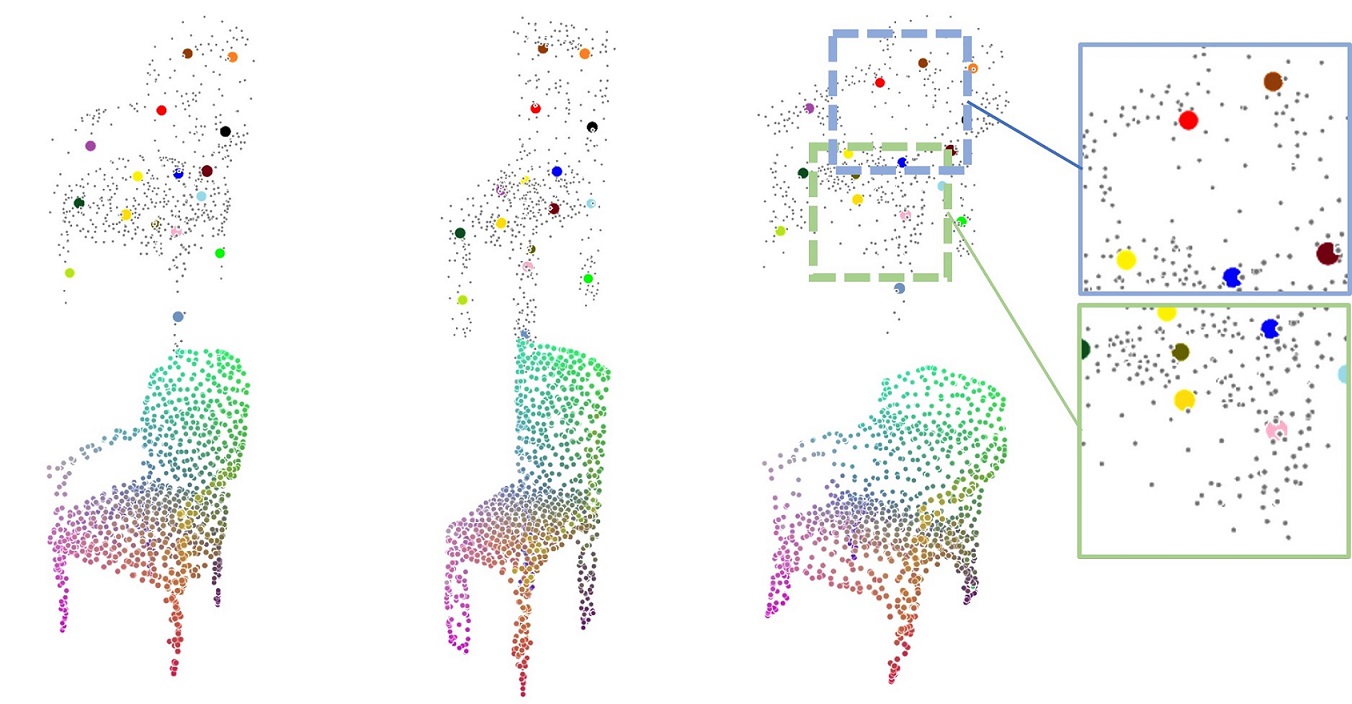}
    \vspace{-2mm}
  \caption{Robustness to non-uniform sampling. The 1st row shows 16 structure points and non-uniform input point clouds. The 2nd row shows 1024 structure points. Corresponding structure points have the same color.}
  \label{fig:diff_sample_non_uniform}
    \vspace{-4mm}
\end{figure}

Table \ref{tab:dif_sample_ave_dis} shows the stability of the structure points with different numbers of uniformly sampled points as inputs. For non-uniformly sampled inputs, the average stability of these categories is $0.3499\%$. We also demonstrate some qualitative results in Figure \ref{fig:diff_sample_vis} and \ref{fig:diff_sample_non_uniform}. As shown by qualitative and quantitative results, our approach is not sensitive to the point sampling density. That is because the PointNet++ \cite{qi2017PointNet++} we used to encode local features is not sensitive to the sampling of point clouds and our point integration module also maintains such property.

\subsection{Testing with Real Scanned Data}
\begin{figure}[htbp]
\centering
  \includegraphics[width=0.47\textwidth]{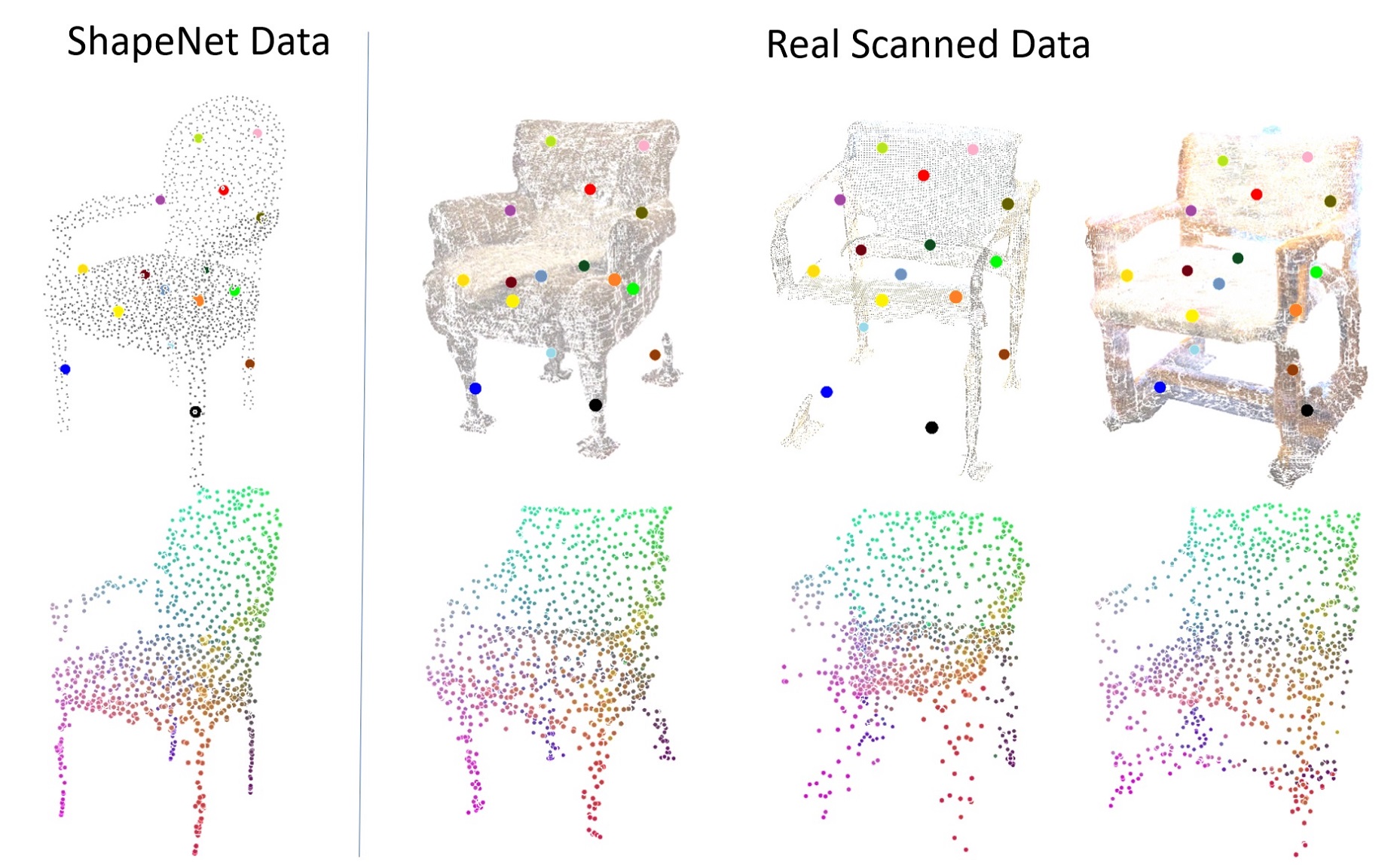}
    \vspace{-2mm}
  \caption{Structure points on real scanned data. We show the produced 16 (1st row) and 1024 (2nd row) structure points for scanned point clouds of chairs. Corresponding structure points are in the same color.}
  \label{fig:vis_realscan}
  
\end{figure}

We show the good generalization of our method by testing on real scanned data. Specifically, we train our network on the ShapeNet dataset with rotation augmentation, and test the trained network on real scanned point clouds \cite{dai2017scannet}. As illustrated in Figure \ref{fig:vis_realscan}, even though the real scanned point clouds are noisy and not seen during training, our network can still produce semantically consistent structure points.

\subsection{Testing with Different Feature Encoder}
\begin{figure}[htbp]
\centering
  \includegraphics[width=0.47\textwidth]{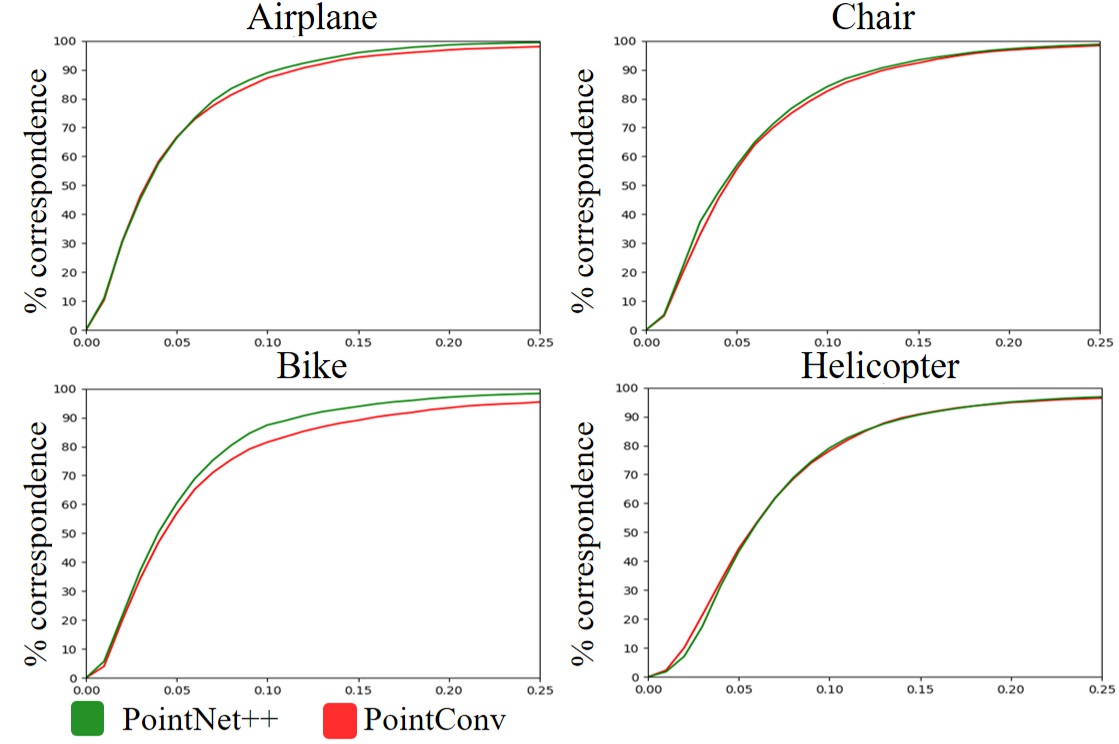}
    \vspace{-2mm}
  \caption{Testing with different point cloud encoder.}
  \label{fig:diff_backbone}

\end{figure}

The proposed point integration module can also be integrated with other point cloud learning architectures to learn consistent structure points. Specifically, We replace the PointNet++ feature encoder with PointConv \cite{wu2019pointconv}, a state-of-the-art point cloud learning architecture, and evaluate the performance of semantic shape correspondence. As shown in Figure ~\ref{fig:diff_backbone}, both architectures can produce similar results on semantic shape correspondence accuracy.

\section{Conclusion}
In this paper, we present a simple yet interpretable unsupervised method for learning a new structural representation in the form of 3D structure points. The produced structure points encode shape structures and exhibit semantic consistency across all the shape instances with similar shape structures.
We evaluate the proposed method with extensive experiments and show state-of-the-art performance on both semantic correspondence and segmentation label transfer tasks. We also show the good generalization of our network by testing on real scanned data. Moreover, our PCA based structure points embedding also has the potential to be used in some important tasks like shape reconstruction and completion, which we would like to explore more in the future.
\paragraph{Acknowledgement} We thank the reviewers for the suggestions, Changjian Li, Guodong Wei, Yumeng Liu for the
valuable discussions.

{\small
\bibliographystyle{ieee_fullname}
\bibliography{ms}
}

\end{document}